# A Methodology for Studying Linguistic and Cultural Change in China, 1900-1950

Spencer Dean Stewart[a]

[a]*School of Information Studies, Purdue University, U.S.A.*

**Abstract**
This paper presents a quantitative approach to studying linguistic and cultural change in China during the first half of the twentieth century, a period that remains understudied in computational humanities research. The dramatic changes in Chinese language and culture during this time call for greater reflection on the tools and methods used for text analysis. This preliminary study offers a framework for analyzing Chinese texts from the late nineteenth and twentieth centuries, demonstrating how established methods such as word counts and word embeddings can provide new historical insights into the complex negotiations between Western modernity and Chinese cultural discourse.

## 1. Introduction

The late nineteenth and early twentieth centuries were a time of significant linguistic and cultural change in China. Existing scholarship has shown that the introduction of thousands of foreign words beginning in the late nineteenth century fundamentally transformed the Chinese language. [1][2] This process was inherently connected to China's pursuit of political, economic, and cultural modernization. In both academic and popular publications, Chinese thinkers explored and debated different paths for China's future as they sought to elevate its position in a changing world. This search for modernity was closely tied to the work of translation, as writers introduced thousands of foreign words into China - a process facilitated by a growing and vibrant print culture that targeted China's sizable literate population (estimated at 90 to 250 million people by the late nineteenth century). [3][4] Subsequently, Chinese elites adopted a new legal and political language within a global framework of empires and assert its territorial claims. [5] The collapse of the Qing in 1911 and the establishment of the Republic of China (1912-1949) created the need for a new language to renegotiate the relationship between the Chinese nation-state and its citizens. [6][7] And efforts toward national integration led intellectuals to push for more colloquial forms of writing and the development of a standardized Mandarin, which could be taught and used throughout the country. [8][9]

This paper provides a preliminary analysis of how computational text analysis methods can be used to study the transformation of Chinese language and culture during the first half of the twentieth century. While the usefulness of methods such as word counts and word embeddings is well established in humanities research, the dramatic shifts in language and culture described above demand a closer reflection on how such changes should inform our approaches to textual

CEUR Workshop Proceedings (CEUR-WS.org)

analysis. At a practical level, these shifts require several additional steps in text preprocessing before the corpus is ready for analysis. This paper is therefore divided into two main sections. The first provides a brief introduction to different textual datasets for studying modern China, an overview of the corpus that will be used in this paper, and a discussion of how the text was prepared for text analysis. Part two shows how different computational methods can be applied to these texts to analyze the evolution of Chinese language and culture. It draws specifically from the framework of *translingual practice* to explore issues of scale and agency in an evolving cultural and linguistic landscape. [1]

## 2. Data and Text Preprossesing

The digitization of Republican China (1912-1949) periodicals, newspapers, and books is still in its infancy, although there has been great progress in recent years. The existing projects have largely been shaped by specific research questions or institutional interests. One such project is The Database for the Study of Modern Chinese Thought and Literature (1830-1930) 中國近現代思想及文學史, currently hosted at National Chengchi University in Taiwan. The team, led by Dr. Cheng Wen-huei, has produced excellent work on the conceptual and intellectual histories of different ideas in late Qing and Republican China. [10] [11] A more recent effort to create a corpus of Republican-era periodicals has been made by the Shanghai Library to digitize and produce full-text editions of their vast collections of periodicals, with a focus on literature, film, and red materials. They have partnered with the University of Chicago's Textual Optics Lab to make portions of the dataset available for analysis through the platform PhiloLogic. [12] Another recent initiative is spearheaded by historian Christian Henriot who has compiled a diverse collection of Modern Chinese texts and made them available for analysis through the platform HistText. [13]

For this preliminary analysis, I have chosen to focus the study on a single periodical of significant cultural significance: *Eastern Miscellany* (東方雜誌). *Eastern Miscellany* was the flagship periodical published by Commercial Press from January 1904 to December 1948. While it was just one of over 20,000 periodicals published during the Republican era, Eastern Miscellany's influence and broad scope make it a useful corpus for research. Historian Ted Hunters, for example, has described Commercial Press as "the most important publisher in China during the first half of the twentieth century," with *Eastern Miscellany* serving as "a key organ of intellectual opinion throughout the forty-some years of its existence." [14] During its print run, *Eastern Miscellany* published on a range of topics such as contemporary events, politics, economics, literature, science, and technology. In total, this corpus includes over 27 thousand individual articles, 5 thousand unique authors, and 96 million total lexical items (see Figure 1). While any dataset is unavoidably imperfect, *Eastern Miscellany* provides an eclectic collection of texts that serves as a useful starting point for future computational humanities research utilizing Republican-era newspapers and periodicals.

The past decade has seen increased attention to the asymmetries involved in multilingual computational text analysis. [15][16] Studies in this realm can be technical in nature, especially when it comes to working with non-Latin languages and scripts. Existing tools such as optical character recognition (OCR) or word segmentation simply don't work as well for languages like Arabic, Japanese, or Chinese. Methods and tools for working with modern Chinese texts have improved considerably in recent years, yet many of these tools are trained to process texts from the last few decades. There are fewer resources and tools for working with textual data

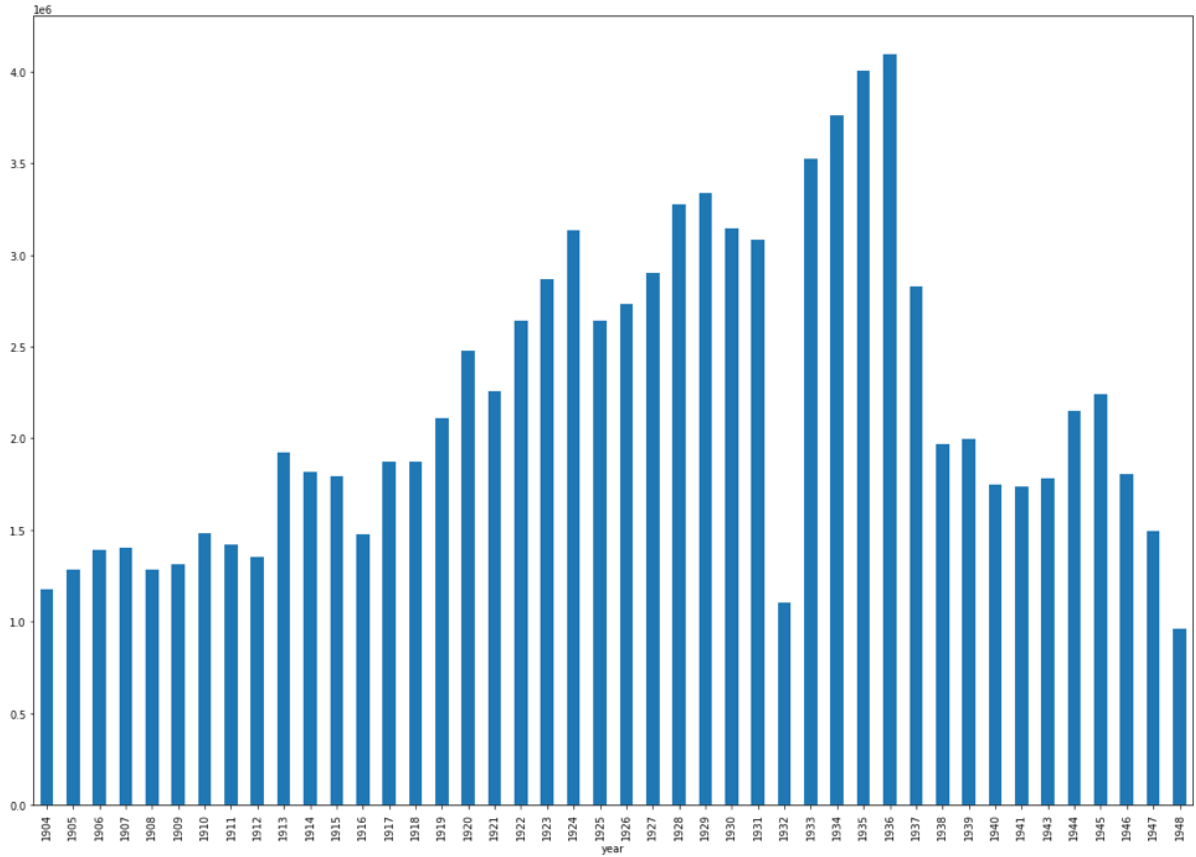

**Figure 1:** Number of Lexical Items by Year (unit: millions)

from the late nineteenth and early twentieth centuries.

The biggest obstacle for working with Chinese texts is word segmentation. After assessing the handful of available tools, we found that Jieba performed best if: (1) we standardized the numerous character variants 異體字; and (2) converted traditional characters to their simplified equivalent. Character variants are Chinese characters that have multiple distinct graphical representations. These variations were commonly used in commercial printing into the twentieth century. The faithful rendering of character variants in the digitization process means that tools such as Jieba are unable to recognize and properly segment common variants. Subsequently, we compiled a list of the 50 most frequently used character variants in our dataset and created a dictionary to convert them into their standardized form. Secondly, we found that while Jieba is capable of processing traditional characters, it performed better when working with simplified Chinese. Therefore, the entire corpus was first converted into simplified Chinese characters prior to conducting text analysis. [12] While such preprocessing steps are not appropriate for all research tasks, they were nevertheless useful in providing a more consistent and standardized corpus to trace cultural change over time.

# 3. Scaling Translingual Practice

In her 1995 book, *Translingual Practice: Literature, National Culture, and Translated Modernity – China, 1900-1937*, Lydia H. Liu argued that scholars need to move beyond conventional influence studies, which might explicitly or implicitly assume a level of universal equivalence when translating concepts and ideas between two languages such as English and Chinese. She instead advocated for studying how new words enter the host language, gain legitimacy, and are creatively employed in popular discourse. Without ignoring the impact that Western modernization discourse had on languages and cultures around the world, this framework allows scholars to more fully explore the intervention of local actors in the process of translating modernity. The idea of translingual practice can shape the way we approach computational text analysis. Rather than simply counting words to measure influence, for example, we need to adopt other methods such as word embeddings to understand how local actors engaged with these new words and concepts. In the case of China, we should avoid assuming equivalence between the imposed language and its Chinese translation. Instead of ending our analysis at the point of adoption, we need to further investigate how Chinese writers debated and engaged with these national and global discourses.

This section employs different methods to study China's linguistic and cultural transformation during the first half of the twentieth century. The first section uses word counts to understand the nature of linguistic change and the word forms that had the greatest impact on Chinese discourse. The second section applies word embeddings to measure cultural and linguistic drift. The final section examines the concept of 'economy' as a case study for how these methods can be used to provide new insights into cultural and linguistic change.

## 3.1. Counting Linguistic and Cultural Change

The easiest way to study the impact of foreign ideas and words on the Chinese language is through word counts. Counting occurrences of foreign words in the pages of *Eastern Miscellany* can provide insights into the nature of language contact and how new concepts shaped the way Chinese writers discussed a variety of topics, from politics and culture to science and law. For this analysis, I compiled the lists of words found in the appendices of *Translingual Practice*, a list that includes around 1,800 different neologisms, loanwords, and transliterations. Figure 2 shows the frequency (count divided by lexical units per year) that these words appeared annually in the pages of *Eastern Miscellany*. It shows that there was a quick increase in frequency during the first two decades of the twentieth century, which stabilized by the 1920s and 1930s (with some exceptions). By the 1920s, these words were on average appearing one to two times per sentence. The most common words included concepts such as government (政府), politics (政治), economy (經濟), and society (社會). They also included ideas about the world (世界) and the international (國際); citizenship (國民), nationality (民族), and voting (選舉); capital (資本), industry (工業), and production (生產); law (法律) and science (科學); revolution (革命) and the military (軍事, 海軍). While many of these conceptions existed in China prior to the twentieth century, much of the language that writers were using to convey ideas about politics, nationalism, economy, law, science, and military was changing.

The act of translating a foreign word into Chinese didn't always lead to its adoption in written discourse. The success or failure of adoption speaks to the nature of language contact during this time. Out of the list of foreign words derived from these appendices, around five hundred don't appear at all in the pages of *Eastern Miscellany*, and half appeared fewer than 50 times.

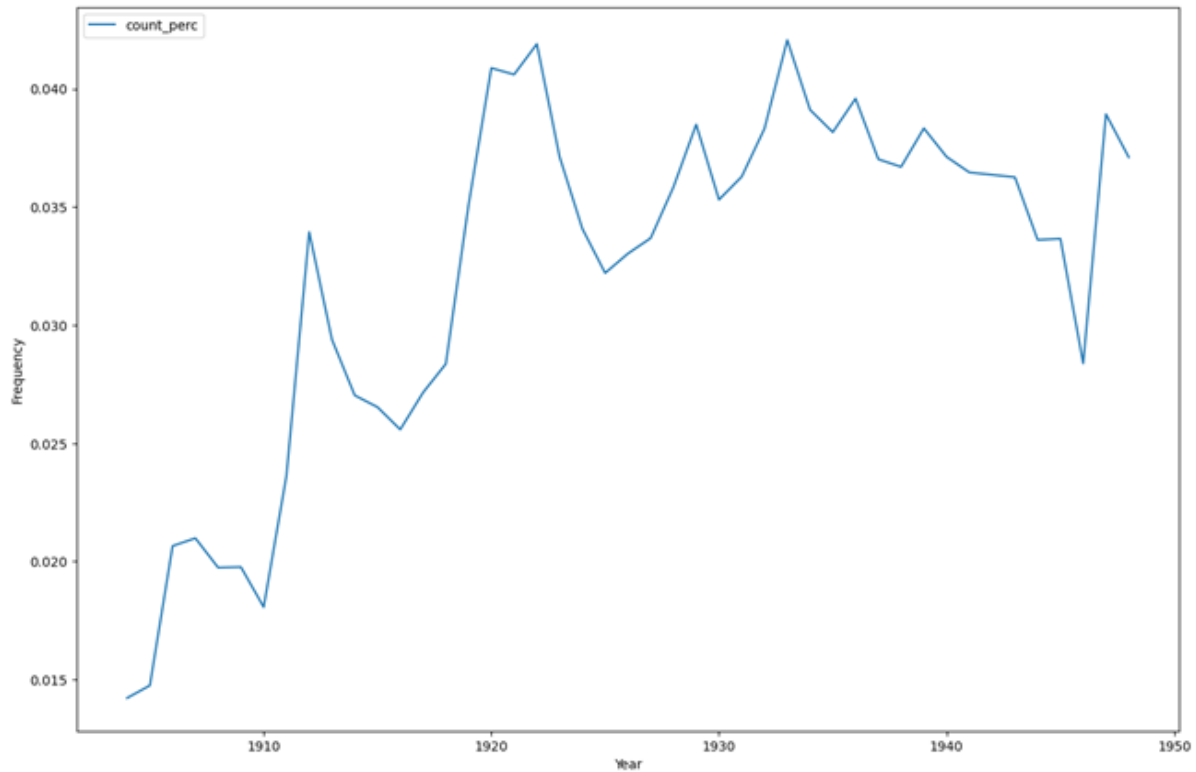

**Figure 2:** Word Frequencies for terms found in *Translingual Practice*, Appendix A, B, C, D, F, G (n=1,788). Frequency is calculated by taking the number of times a word appeared divided by total lexical items per year.

There is a strong relationship between rates of adoption and the mode of introduction. Liu divided newly introduced terms into several different categories. The first and earliest of these were neologisms derived from Missionary-Chinese texts (Appendix A; n=184). These included a variety of terms such as bread (麵包), newspaper (報紙), physics (物理) and university (大學). The next series of words arrived in China by way of Japan. First among these were terms coined by Japanese translators who used Chinese characters to translate European words that were later introduced into China (Appendix B; n=489). These included words such as biology (生物學), comic books (漫畫), debt (債務), factory (工廠), international law (國際公法), the natural sciences (自然科學), and voting (投票). A second, smaller group of words, consisted of *kanji* that arrived in China without necessarily involving European languages (Appendix C; n=51), such as rickshaw (人力車) and religion (宗教).

The third and most important type was *return graphic loans*. These were Chinese-character compounds with established meanings in classical Chinese, which Japanese translators adapted to correspond to modern European words (Appendix D; n=234). Some of these can be viewed as extensions of existing words whose meaning in some way resembled the foreign equivalent. For example, the character pair *geming* (革命) traditionally referred to dynastic transition, but in the late nineteenth and twentieth centuries it came to mean *revolution*. At other times, translators used existing character pairs to introduce entirely different ideas. The character pair *jiantao* (檢討), discussed more below, went from being a title in the Hanlin academy to meaning something approximating *examination* or *self-criticism*. The final and largest number of new terms consisted of transliterations from English, French, and German (Appendix F; n=761), and transliterations from Russian (Appendix G; n=78). Some of these transliterations are recognizable today, such as logic (邏輯), microphone (麥克風), or Darwinism (達爾文主義), but most have fallen out of use (e.g., communist 康門尼斯特 and syrup 舍利別).

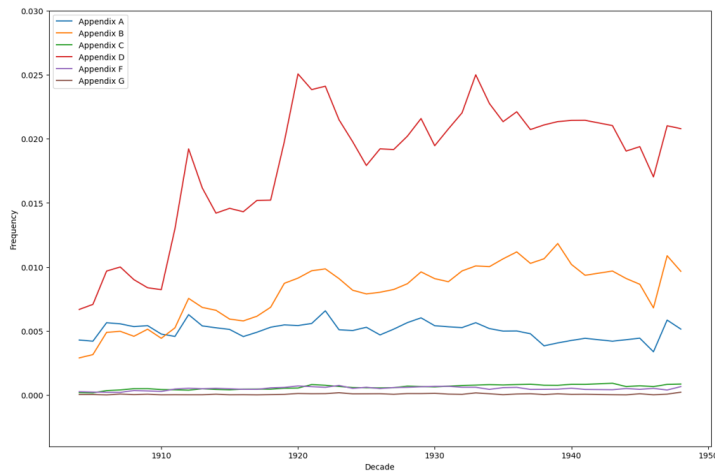

**Figure 3:** Frequency by appendix. Frequency is calculated by taking the number of times the words appear in each appendix divided by total lexical items per year.

Based on total counts, one might assume that transliterations were the most common and possibly most influential form of language contact. However, frequency counts based on these categorizations confirms what existing scholarship has said about the importance of Japanese translators in shaping Chinese written discourse. As shown in Figure 3, the most frequently appearing words were return graphic loans (Appendix D). The most common of these included words recognizable today, such as government (政府), society (社會), economy (經濟), politics (政治), and world (世界). These loanwords, which at times were just extensions of existing words, were apparently more easily and more readily adopted by Chinese writers. Moreover, if we look at frequency per word rather than frequency by appendix, we find that return graphic loans were nearly three times more likely to appear than words derived from Missionary-Chinese texts (Appendix A). Terms coined by Japanese translators using Chinese characters (Appendix B) also increased, but not as frequently as return graphic loans. Perhaps most interesting were that transliterations, despite constituting the largest total number of overall neologisms, ranked low on the list compared to other appendices. There are clear examples of transliterations from this period that became localized (see opium 鴉片, coffee 咖啡, logic 邏輯 or Soviet 蘇維埃), but most transliterations weren't adopted. Overall, return graphic loans were more than 46 times more likely to appear in the pages of *Eastern Miscellany* than transliterations (see Table 1).

### 3.2. Measuring Change

Word frequencies can only tell us so much about Sino-Foreign interactions. A more *China-centered* approach doesn't require that we dismiss the impact of the West in shaping modern China; rather, it emphasizes the active participation of Chinese actors in this process. [17] Is it possible to computationally capture the agency of Chinese writers in translating modernity? Can we adopt a form of distant reading informed by the concept of translingual practice?

**Table 1**
Words and Word Counts by Appendix

| Appendix | Num. of Words | Total Count | Average | Most Frequent Words |
|---|---|---|---|---|
| A | 184 | 492,949 | 2,664.5 | 會議; 委員; 雜誌; 法律; 選舉; 鐵路; 學校; 帝國 |
| B | 489 | 824,413 | 1,679.0 | 國際; 政策; 銀行; 民族; 工業; 財政; 目的; 承認 |
| C | 51 | 62,469 | 1,224.8 | 宗教; 表現; 內容; 手續; 方針; 集團; 距離; 目標 |
| D | 234 | 1,843,495 | 7,844.7 | 政府; 社會; 主義; 經濟; 方面; 政治; 世界; 關係 |
| F | 761 | 51,443 | 67.4 | 馬克; 蘇維埃; 鴉片; 基督教 |
| G | 78 | 7,932 | 101.7 | 蘇維埃; 盧布; 烏拉 |

Temporal word embeddings are particularly well suited for studying history, especially when it comes to understanding linguistic, cultural, and conceptual change. Word embeddings build on J.R. Firth's concept that "you shall know a word by the company it keeps." We can conclude that words such as dog and bone or cat and mouse are related given that they often appear near each other in a sentence. Word embeddings take this logic one step further by looking for words that share the same co-occurring terms, thereby clustering words that have similar semantic meanings. This might include synonyms or words that share specific conceptual relationships such as animals, countries, or automobile companies. One way to think about this method is that word embedding models place words within a multidimensional space where the distance between individual words represents how closely they are conceptually or linguistically related. Previous scholarship has demonstrated the utility of this method in measuring cultural and linguistic change over time. [18][19] Scholars have used temporal word embeddings to propose laws of semantic change, [20] uncover racial inequality in postwar U.S. fiction, [21] discover the invisible labor of women's editorial work, [22] highlight representations of race and ethnicity under the Japanese empire, [23] and reconstruct networks of influence in abolitionist writings. [24]

Here I use temporal word embeddings to provide qualitative insights into the larger systematic changes measured above. To do this, I created separate word embedding models for distinct periods of time. I experimented with different periodizations, including those that conformed to conventional political events, but for simplicity, I divided the corpus by decade (i.e., 1904-1909; 1910-1919; 1920-1929; 1930-1939; 1940-1949). After aligning the models, [20] I then measured (using cosine similarity scores) how a given word was or was not changing over time .

A useful example is the character pair *jiantao* 檢討, which underwent significant change during the first half of the twentieth century. As mentioned above, *jiantao* originally referred to a title/name associated with the Hanlin academy. In Japan, this two-character compound was used to translate the concepts of examination and self-criticism. *Jiantao* as *examination* was then introduced into China in the early twentieth century. Figure 4 provides a two-dimensional representation of the character pair as its meaning changed in the pages of *Eastern Miscellany*. Here we find it retaining its original meaning during the first decade of the twentieth century and gradually shifting to take on the meaning of examination by the 1940s.

*Jiantao* is just one example of how word embeddings can be useful for quantifying how return graphic loans transformed existing written language. Using cosine similarity scores, we can further isolate words that experienced a noticeable shift during the first half of the twentieth century. As shown in Table 2, such examples include the character pair *youji* 游擊, which evolved from its traditional usage as a military position to refer to *guerrilla warfare*. *Jingji* 經

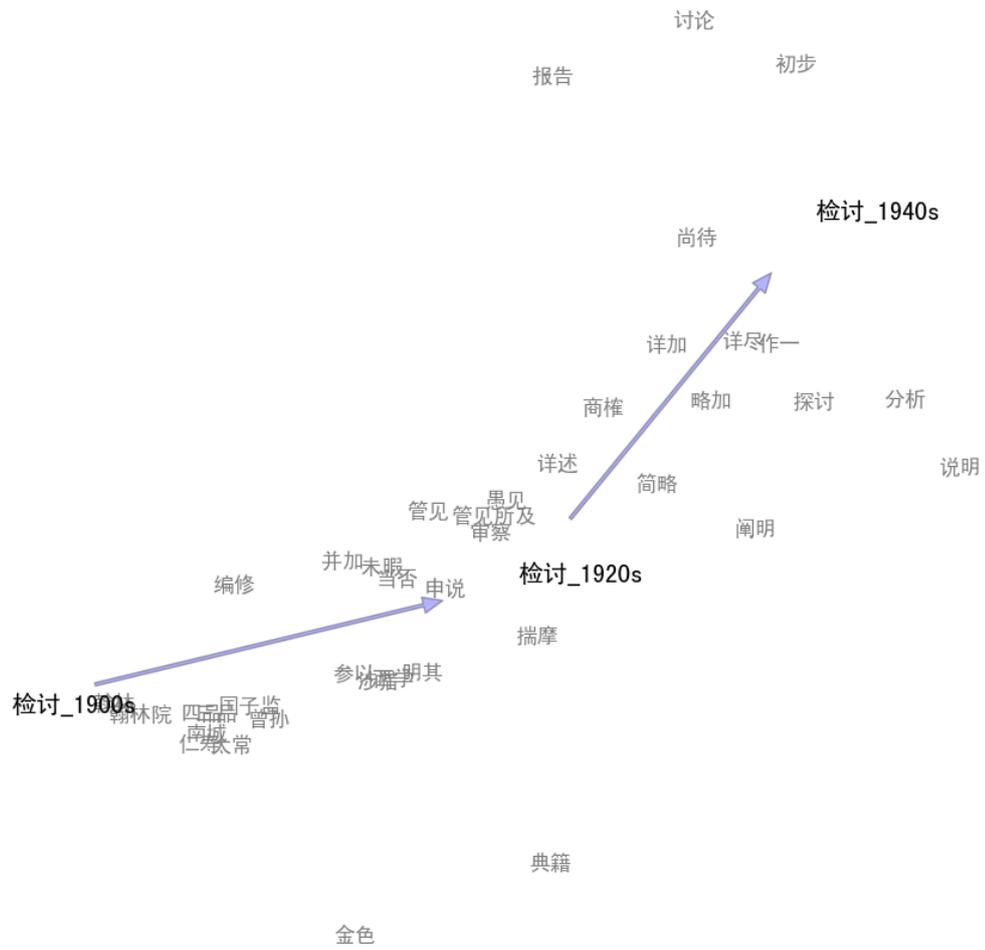

**Figure 4:** Graphical representation of the semantic change to *jiantao*, 1900s-1940s.

濟, as discussed more below, once referred broadly to political governance before it was used to translate economy. The characters *fengjian* 封建 were used to translate the idea of feudal and feudalism, despite the fact that this character pair was long used by famous thinkers for around two thousand years to refer to the granting (feng) of land to establish (jian) vassals. [25] [26] Finally, the characters *geming* 革命 once referred to dynastic transition, but in the late nineteenth and twentieth centuries took on the more active meaning of revolution.

**Table 2**
Examples of Semantic Drift for Return Graphic Loans from 1904-1909 to 1940-1948

| Word | Pinyin | Original Meaning | New Meaning | Cosine Similarity |
|------|--------|------------------|-------------|-------------------|
| 檢討 | jiantao | Title-Hanlin Academy | Examination, self-criticism | 0.107032 |
| 游擊 | youji | Military position | Guerrilla warfare | 0.391787 |
| 經濟 | jingji | Political Governance | Economy | 0.522084 |
| 封建 | fengjian | Offering and establishing | Feudal | 0.537265 |
| 革命 | geming | Dynastic change | Revolution | 0.638383 |
| 教授 | jiaoshou | Impart knowledge | Professor | 0.748040 |

### 3.3. Economy in the Popular Press

The mapping of the word economy onto the Chinese character pair *jingji* (經濟) involved complicated interactions between China, Japan, and the West. As Tianyu Feng has shown, *jingji* was a term that originally meant governance for the people. It included aspects that would later be identified as economy, such as finance, trade, and transportation. But even in the late Qing, these aspects were subsumed under the umbrella of politics. In many ways, this was comparable to how the word economy was used in Western discourse prior to the eighteenth century. It therefore makes sense that early Japanese writers originally chose to translate it as *jingji* (or *keizai*). However, the suitability of this translation began to be called into question as European thinkers and classical economists constructed an idea of economy and economics that was distinct from politics and political economy, to instead focus more narrowly on production, distribution, and consumption. [25]

The evolving meaning of economy in the West posed problems for Chinese translators in the late nineteenth century. Influential figures such as Liang Qichao and Yan Fu avoided *jingji* as a translation for economy, instead preferring other translations such as *fuguoce* 富國策 (strategy of enriching the country), *cailixue* 財理學 (financial management), *licaixue* 理財學 (wealth administration), *zisheng* 資生 (welfare or wellbeing), *shengjixue* 生計學 (study of livelihood), and *jixue* 計學 (study of counting), along with several different transliterations (e.g., 葉科諾密, 愛康諾米, and 依康老米). Most Japanese writers continued to use the term *jingji/keizai* to translate economy, with some Chinese commentators suggesting that these writers mistranslated the word by mistaking two Chinese characters sharing the same pronunciation. But over time, the influence of *jingji* won out as prominent political figures such as Sun Yat-sen helped to solidify it as the accepted translation of economy. [25][27]

Word frequencies reveal the contested nature of different translations during the first decade of the twentieth century, along with the eventual rise of *jingji* as the preferred translation. The only real contenders to *jingji* in the pages of *Eastern Miscellany* were *jixue* (roughly 145 appearances) and *licaixue* (around 68 appearances), with both of these surpassing *jingji xue* 經濟學 (economics) for parts of the first two decades of the twentieth century. In the 1900s and 1910s, writers used these different translations to appeal to the authority of economists (*jixue jia* 計學家), propose the establishment of centers for economic study (*jixue guan* 計學館), and provide introductions to economic theories and axioms. However, by the early 1910s, the term *jingji* and *jingji xue* to refer to the economy and the study of economics, respectively, gained prominence over these alternatives. Overall, in the pages of *Eastern Miscellany*, the term *jingji xue* (economics) appeared around 2,900 times and *jingji* (economy) over 58,000 times.

Amidst debates regarding the proper translation for economy, Chinese writers explored the meaning of *jingji* as a concept and its role in Chinese politics and society. Chinese thinkers had long written broadly about what would come to be called the economy in the twentieth century. As Margareta Zanasi shows, the commercial revolution of the mid-1500s led Chinese thinkers to adopt pro-market and pro-consumption ideas. By the 1800s, just as the West was beginning to embrace these same ideas, Chinese officials and intellectuals were faced with new internal and external circumstances that led them to instead promote "a hybrid form of market economy with ad hoc government intervention." This system "more closely resembled forms of developmental state and was intended to solve the China-specific threats of scarcity and, after the mid-1800s, of imperialist interference." [28] The language that they used to discuss the economy was therefore changing just as Chinese economic thinkers moved away from emphasizing stability to stressing the importance of economic growth for nation building. According to Zanasi, the Chinese economic crisis "came to be framed in terms of the struggle of the Chinese race and civilization for the survival of the fittest." [28]

A distant reading of the term *jingji* in the pages of *Eastern Miscellany* provides several interesting insights into the nature of economic discourse within China. First, the idea of economy moved from a more abstract notion during the first decade of the twentieth century to focus on concrete industries and sectors. As shown in Table 3, terms closely associated with *jingji* during the first decade were relatively abstract, suggesting that the economy was understood more as a *concept* and *phenomenon*. The place of social Darwinism within this discourse is also suggested by its close association with *evolution*. Different articles drew from a social Darwinian framework to discuss social and economic progress in China, including the (im)possibility of eliminating socioeconomic inequality, the advancement of commercial interests, and details of monetary policies. The idea of *rise and fall* in relation to the economy also existed in the first two decades of the twentieth century. At times, it was used to discuss the rise and fall of the national economy (國民經濟之盛衰). At other times, it referred to trade, industry, companies, and even urban centers. But overall, the general trend by the 1910s pointed in the direction of a more concrete understanding of what constituted the economy. Commerce, industry, international interactions, finance and trade, along with the creation of businesses (e.g., 經營締造) were closely aligned with this new conceptualization of what constituted *jingji*.

A second observation points to an industrial and production bias within popular economic discourse. Zanasi has argued that although economic commentators recognized that consumption was not inherently bad, their focus was often on how the overall economy failed to move forward in "purchasing power and consumption habits." [29] This was caused by gaps in levels of development within China, where urban populations in Shanghai could enjoy luxurious lifestyles while rural populations lacked basic necessities. At first glance, Table 3 suggests that writers in the pages of *Eastern Miscellany* shared this concern as *jingji* was closely associated with industry and industrial production (e.g., 產業, 工業, and 實業). Even in these vernacular discussions of economy aimed at an urban audience, appearances of *production* (生產, n=25,858) occurred nearly four times more often than *consumption* (消費, n=6,528), and over ten thousand more times than *commerce* (商業, n=14,088). References to the industrial sector (工業, n=25,945) also appeared nearly twice as often as agriculture (農業, n=13,819). Cotton provides an interesting example where agriculture and industry met as the cotton industry became more closely associated with *jingji* in the 1930s and 1940s. Situated between the rural economy and urban industry, cotton was an important strategic sector promoted by the government in the 1930s and into the 1940s. [30] Discussions of the rural economy, agriculture, and consumption were frequent enough. But industrial production was clearly the priority for much of this time.

**Table 3**
Selected words closely associated with *jingji* by decade

| 1904-1909 | 1910-1919 | 1920-1929 | 1930-1939 | 1940-1949 |
|---|---|---|---|---|
| Concept 觀念 | Higher level 上及 | Productive Sector 產業 | Productive Sector 產業 | Finance 金融 |
| Rise & Fall 盛衰 | Commerce 商業 | Commerce 商業 | Commerce 商業 | Public Finance 財政 |
| Thought 思想 | Industry/Commerce 工商業 | Industry/Commerce 工商業 | Finance 金融 | Cotton Industry 棉業 |
| Function/Effect 作用 | National Defense 國防 | Industrial Sector 工業 | Industry/Commerce 工商業 | Industry/Commerce 工商業 |
| Evolution 進化 | Foreign 對外 | Agriculture 農業 | Industry 實業 | Commerce 商業 |
| Beneficial To 有益於 | Rise & Fall 盛衰 | Finance 金融 | Public Finance 財政 | Currency System 幣制 |
| Phenomenon 現象 | Found/Create 締造 | Command/Control 統制 | Cotton Industry 棉業 | Hub 樞紐 |

A third finding relates to the occurrence of the word *control* (統制) in the 1920s, the appearance of which serves as a cautionary tale regarding the need to balance distant and close reading. The idea of a control economy in Republican China is most closely associated with the 1930s and 1940s when the government promoted Fascist models of command economy. [29] Its appearance in the 1920s is therefore notable and at first glance might suggest popular discussions of this form of economic governance prior to its implementation in the 1930s. Upon closer reading, we find that the correlation between control and economy in the 1920s derives from a handful of articles describing foreign economies such as Germany and Japan, along with warnings about the potential dangers of government control of industry. Rather than foreshadowing developments in the 1930s, the appearance of *control* in the 1920s more accurately represents how some writers were responding to global economic discourse and expressing reservations about government oversight. The relationship between economy and control beginning in the 1930s is better understood through the appearance of cotton (棉業) in the 1930s as *Eastern Miscellany* occasionally reported on the Cotton Control Commission (棉業統制委員會), an organization established by the Nationalist government in 1934 as a way to directly control the circulation of raw cotton and the production of cotton goods. [29][31] The appearance of control in the 1920s therefore illustrates the need to balance close reading with distant reading. It also demonstrates how Chinese writers were actively observing and commenting on global economic affairs.

This preliminary conceptual history of *jingji* as found in the pages of *Eastern Miscellany* demonstrates how Chinese writers were influenced by and contributed to Western economic discourse. They debated the appropriateness of different translations and adjusted their writings in accordance with changing local and global developments. Distant reading provides one way to look at this process. The arrival of thousands of new words and ideas clearly transformed the Chinese language. At the same time, the significance of these concepts and their meaning for the Chinese condition were constantly being renegotiated.

# 4. Conclusion

China's linguistic and cultural transformation that began in the late nineteenth century serves as a valuable case study for applying computational methods to historical data. These language changes require several additional steps in the text preprocessing stage to create a standardized corpus capable of more accurately capturing cultural change. The scale of this transformation also provides opportunities for computational research. As shown in this paper, word frequencies shed light on the nature of language contact in China during the first half of the twentieth century as foreign words were introduced from the West via Japan, with new concepts often being mapped onto older words. Word embeddings shift our attention towards how these words were used within Chinese written discourse, including the repurposing of its linguistic past to present a more international and modern present. Through a combination of distant and close reading of the term economy (*jingji*), this paper also demonstrates how Chinese writers responded to global discourse. Chinese writers introduced, negotiated, rejected, debated, and finally employed new concepts in creative and interesting ways. This study suggests that computational humanities research, when combined with close attention to historical and cultural context, can deepen our understanding of the interactions between method, language, translation, and culture.